\documentclass[conference]{IEEEtran}

\makeatletter

\def\ps@IEEEtitlepagestyle{%
  \def\@evenfoot{}%
}

\usepackage{eso-pic}
\IEEEoverridecommandlockouts
\usepackage{cite}
\usepackage{amsmath,amssymb,amsfonts}
\usepackage{algorithmic}
\usepackage{algorithm}
\usepackage{graphicx}
\usepackage{textcomp}
\usepackage{url}
\usepackage{xcolor}
\def\BibTeX{{\rm B\kern-.05em{\sc i\kern-.025em b}\kern-.08em
    T\kern-.1667em\lower.7ex\hbox{E}\kern-.125emX}}

\usepackage{eso-pic}
\newcommand\AtPageUpperMyright[1]{\AtPageUpperLeft{%
 \put(\LenToUnit{0.17\paperwidth},\LenToUnit{-2cm}){%
     \parbox{0.9\textwidth}{\raggedleft\fontsize{8}{11}\selectfont #1}}%
 }}%
\newcommand{\conf}[1]{%
\AddToShipoutPictureBG*{%
\AtPageUpperMyright{#1}
}
}

\begin{document}
\title{\vspace*{1cm} Real-TurnTurk: A Multimodal Turkish Corpus for Turn-Taking Prediction
\\
}

\author{\IEEEauthorblockN{Ahmet Tuğrul Bayrak}
\IEEEauthorblockA{\textit{Data Science and Innovation} \\
\textit{Ata Technology Platforms}\\
Istanbul, Turkey \\
tugrul.bayrak@atptech.com}
\and
\IEEEauthorblockN{Fatma Nur Korkmaz}
\IEEEauthorblockA{\textit{Data Science and Innovation} \\
\textit{Ata Technology Platforms}\\
Istanbul, Turkey \\
fatmanur.korkmaz@atptech.com}
\and
\IEEEauthorblockN{Bekir Berker Türker}
\IEEEauthorblockA{\textit{Data Science and Innovation} \\
\textit{Ata Technology Platforms}\\
Istanbul, Turkey \\
berker.turker@atptech.com}
\and[\hfill\mbox{}\linebreak\mbox{}\hfill]
\IEEEauthorblockN{Mustafa Sertaç Türkel}
\IEEEauthorblockA{\textit{Data Science and Innovation} \\
\textit{Ata Technology Platforms}\\
Istanbul, Turkey \\
sertac.turkel@atptech.com}
\and
\IEEEauthorblockN{Alper Kaplan}
\IEEEauthorblockA
{
\textit{Independent Researcher}\\
Luxembourg, Luxembourg \\
alper.kaplan@outlook.com}

}

\maketitle
\conf{\textit{ 6. Interdisciplinary Conference on Electrics and Computer (INTCEC 2026) \\
24-25 September 2026, Chicago-USA}}

\begin{center}
\small
Accepted to INTCEC 2026. This is the author's pre-print version. The final authenticated version will be available through the conference proceedings.
\end{center}

\begin{abstract}

Turn-taking is a basic organizational feature of human conversation and remains difficult to model in natural, synchronous dialog systems. While existing research has explored multimodal approaches and large language models for turn-ending prediction, there is a lack of naturalistic conversational corpora specifically addressing turn-taking dynamics in Turkish. This study introduces a multimodal Turkish conversational dataset of unscripted dyadic interactions, comprising synchronized front-facing video, per-speaker audio channels that allow overlapping speech to be attributed to individual speakers, and time-aligned transcriptions. Turn-taking prediction is formulated as a binary classification problem, and a Genetic Algorithm (GA) is employed to optimize interpretable decision rules derived from visual, acoustic, and linguistic features. A hybrid AND-OR rule representation is adopted in the proposed framework to represent the alternative cue combinations that precede a turn transition. 
\end{abstract}


\section{Introduction}

Spoken dialogue systems built on Large Language Models (LLMs) are now widely deployed, however several interaction-level problems remain unresolved. Many current systems rely on silence detection to determine turn completion: the system assumes the user has finished speaking after a predetermined period of silence. The core limitation is the variability of human speech: some speakers proceed rapidly with minimal pauses, while others require longer intervals to formulate their thoughts without yielding the turn. When a system misjudges these cues and transitions prematurely, speech overlaps occur and dialogue synchronization degrades.

Sacks et al. \cite{sacks1974} introduced turn-constructional units (TCUs) and transition relevance places (TRPs), a framework that still guides computational and empirical work. Silence durations and turn offsets carry paralinguistic meaning and reflect individual conversational styles, and simple acoustic thresholds cannot reliably capture the hold/shift distinction \cite{patamia2025, threlkeld2022}; conversational context and task type further alter turn-taking dynamics, indicating that metrics derived from structured elicitation tasks may not generalize to naturalistic dialogue \cite{watson2020}.

In computational modeling, standard metrics often fail to capture the real-time trade-off between response latency and false cut-ins \cite{lala2018}, and reviews stress that the absence of standardized multilingual benchmarks remains a significant limitation \cite{skantze2021, castillo2025}. Continuous, frame-level formulations have increasingly been used in place of silence thresholds: multiscale recurrent models predict upcoming speech activity from multimodal streams \cite{roddy2018}, and Voice Activity Projection learns turn-taking events self-supervised from future voice activity, with real-time variants now embedded in incremental dialogue systems \cite{ekstedt2022, inoue2024}; full-duplex speech foundation models integrate turn management into the generative model itself \cite{defossez2024}. While language models and lexico-syntactic features improve turn-ending prediction in text settings \cite{razavi2021, pinto2024}, multimodal fusion of audio, text, and gesture is reported to improve on unimodal baselines \cite{pinto2024, lin2025}. Social dynamics such as participatory profiles and user competency further affect conversational equity, which motivates adaptive models \cite{hu2022, chattaraman2019}.

Turkish conversational resources have expanded to sentiment analysis and synthetic turn-taking datasets \cite{polat2024, bayrak2026}; however, a gap remains for naturalistic Turkish corpora with primary turn-taking annotations. This study introduces such a corpus of unscripted dyadic interactions, with synchronized video, per-speaker audio channels in most interactions, and time-aligned transcriptions, and optimizes rules over features from each modality with a genetic algorithm to detect turn change points.

\section{Data}

The dataset comprises 11 unscripted dyadic interactions with 4.23 hours (253.6 min) of synchronized VTT text, WebM audio, and MP4 video. The interactions were recorded with one 48\,kHz WebM file per speaker, which allows overlapping speech to be attributed reliably to a single speaker. Video is 1920$\times$1080 at 16\,fps, with the two participants occupying the left and right halves of the frame. The transcripts contain 5{,}383 speech segments and 35{,}728 words with millisecond-precision timestamps and speaker labels.

Of the total duration, 222.1\,min (87.6\%) is speech and 31.5\,min (12.4\%) is non-speech. Because overlapping speech is counted for both participants, the summed speaker talk time is 231.4\,min; this is the reference total against which the percentages of Table \ref{speaker_cont} are computed, whereas the word percentages of the same table are computed against the 35{,}728-word total. Under the criteria of Section \ref{sec:labelling}, 1{,}750 turn-taking instances were annotated, excluding backchannels from positives, against 3{,}500 sampled negatives. Per-conversation statistics are listed in Table \ref{dataset_stats} and per-speaker statistics in Table \ref{speaker_cont}.

The speakers are not evenly represented. As the \textit{\# Conv.} column of Table \ref{speaker_cont} shows, two act as recurring interlocutors: user\_01 in 5 conversations and user\_03 in 6. These two never converse with each other, and every conversation contains exactly one of them; together they account for 60.8\% of words and 60.1\% of speaking time. Of the rest, user\_04 and user\_06 appear twice and the others once. This recording design, in which two speakers act as recurring interlocutors, produces the speaker-level imbalance and directly constrains the cross-validation protocol (Section \ref{sec:cv}). A subset of the dataset is available on Hugging Face at \url{https://huggingface.co/datasets/tugrulbayrak/Real-TurnTurk}.

\begin{table}[t]
\caption{Statistics by Conversation}
\begin{center}
\setlength{\tabcolsep}{4pt}
\begin{tabular}{|l|c|c|c|c|c|}
\hline
\textbf{Speaker Set} &
\textbf{\begin{tabular}[c]{@{}c@{}}Dur.\\ (min)\end{tabular}} &
\textbf{\begin{tabular}[c]{@{}c@{}}Speech\\ (min)\end{tabular}} &
\textbf{\begin{tabular}[c]{@{}c@{}}Non-sp.\\ (min)\end{tabular}} &
\textbf{\begin{tabular}[c]{@{}c@{}}Non-sp.\\ (\%)\end{tabular}} &
\textbf{\# Words} \\
\hline
\textit{user\_01 \& user\_02} & 26.5 & 22.7 & 3.7 & 14.1 & 3{,}267 \\
\hline
\textit{user\_03 \& user\_04} & 16.0 & 13.7 & 2.3 & 14.4 & 2{,}236 \\
\hline
\textit{user\_01 \& user\_05} & 16.4 & 13.6 & 2.8 & 17.2 & 2{,}219 \\
\hline
\textit{user\_01 \& user\_04} & 24.4 & 20.9 & 3.5 & 14.5 & 3{,}290 \\
\hline
\textit{user\_03 \& user\_06} & 17.1 & 14.1 & 3.1 & 17.9 & 2{,}252 \\
\hline
\textit{user\_03 \& user\_07} & 25.5 & 22.9 & 2.6 & 10.1 & 3{,}856 \\
\hline
\textit{user\_03 \& user\_08} & 28.5 & 24.6 & 3.9 & 13.7 & 3{,}960 \\
\hline
\textit{user\_01 \& user\_06} & 17.1 & 14.5 & 2.5 & 14.8 & 2{,}174 \\
\hline
\textit{user\_03 \& user\_09} & 25.1 & 23.1 & 2.0 & 8.1 & 3{,}917 \\
\hline
\textit{user\_03 \& user\_10} & 25.7 & 23.0 & 2.7 & 10.5 & 3{,}952 \\
\hline
\textit{user\_01 \& user\_11} & 31.4 & 29.0 & 2.4 & 7.6 & 4{,}605 \\
\hline
\textbf{Total} & \textbf{253.6} & \textbf{222.1} & \textbf{31.5} & \textbf{12.4} & \textbf{35{,}728} \\
\hline
\end{tabular}
\label{dataset_stats}
\end{center}
\end{table}

\subsection{Turn Labelling}
\label{sec:labelling}

\begin{figure}[t]\centerline{\includegraphics[width=0.72\columnwidth]{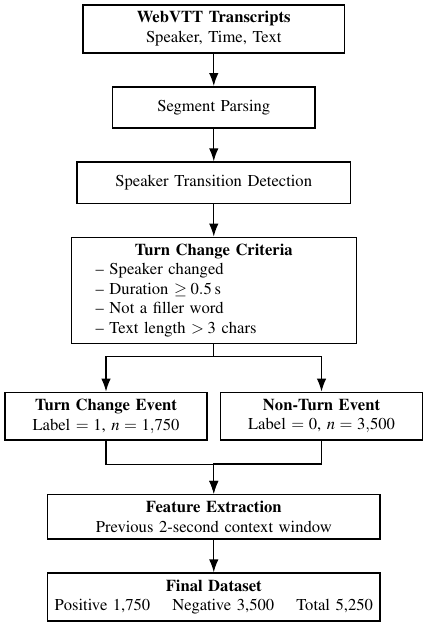}}
\caption{Turn labelling flow.}
\label{labeling_flow}
\end{figure}

A turn change is the moment when the current speaker ends and the next begins. Turn changes are identified by a semi-automatic procedure over VTT transcripts: parsing cues, merging consecutive cues that carry the same speaker label into a single continuous turn, detecting speaker transitions between merged turns, and applying a rule-based filter that separates genuine turns from backchannels. Merging is necessary because sequential cues from one speaker are a segmentation artifact and must not be counted as transitions. As shown in Fig. \ref{labeling_flow}, a transition is a turn change only when four criteria hold concurrently:

\begin{equation}
\label{eq:turnchange}
\text{TurnChange}(t) = 1 \iff
\begin{aligned}
& (S_i \neq S_{i+1}) \\
& \land (d_{i+1} \geq 0.5s) \\
& \land (T_{i+1} \notin \text{FillerWords}) \\
& \land (\text{len}(T_{i+1}) > 3)
\end{aligned}
\end{equation}

Where $t$ denotes the onset time of segment $i+1$; $S_i$ and $S_{i+1}$ represent the speakers of the current and subsequent segments; $d_{i+1}$ is the duration of the subsequent segment; and $T_{i+1}$ is its textual content, with $\text{len}(\cdot)$ measured in characters.

\begin{itemize}\itemsep0pt
    \item \textit{Speaker}: the subsequent segment must come from a different speaker.
    \item \textit{Duration}: a 0.5\,s threshold is applied, since minimal feedback signals in Turkish (e.g., ``hıhı'', ``mm'') typically last under 0.5\,s while genuine turn initiations are longer.
    \item \textit{Content}: segments whose text is a Turkish filler token, or whose surface form is three characters or fewer, are excluded as acknowledgments.
    \item \textit{Temporal}: the timestamp is fixed at the onset of the incoming speaker's segment, giving a precise reference point for feature extraction.
\end{itemize}

Two labelling criteria (duration and filler status) overlap with linguistic features (\textit{word\_duration}, \textit{is\_filler}) available to the optimizer, which could allow a rule to reproduce the labelling function rather than conversational structure. The two are computed on disjoint speech, however: \eqref{eq:turnchange} is evaluated on the \emph{incoming} segment $i+1$, whereas features cover the 2.0\,s window \emph{preceding} $t$, i.e.\ the outgoing speaker. This is also why \textit{is\_filler} can act as a positive predictor although fillers are excluded from positive labels: the occurrences refer to different speakers.

3{,}500 negatives were chosen uniformly at random from all points that satisfy the feature-extraction requirements yet are not annotated turn changes, with a $\pm 1.0$\,s exclusion buffer around every positive, ensuring that adjacent windows cannot leak across classes. Sampling was stratified per conversation in proportion to duration, seed 42. Backchannel onsets are excluded from positives by \eqref{eq:turnchange}; they are eligible as negatives. Because 12.4\% of the corpus is non-speech, most negatives fall inside ongoing speech rather than in pauses. The 1:2 ratio was fixed a priori and not tuned.

Overlapping speech is not excluded from the analysis. 81.4\% of annotated transitions contain some overlap, measurable per speaker in the nine interactions with separated channels.

\begin{table}[b]
\caption{Speaker Contribution Statistics}
\begin{center}
\setlength{\tabcolsep}{3.5pt}
\begin{tabular}{|l|c|c|c|c|c|}
\hline
\textbf{Speaker}  & \textbf{\# Words} & \textbf{\% Words} & \textbf{\begin{tabular}[c]{@{}c@{}}Speak.\\ (min)\end{tabular}} & \textbf{\begin{tabular}[c]{@{}c@{}}\% Speak.\\ time\end{tabular}} & \textbf{\# Conv.} \\
\hline
\textit{user\_01} & 9{,}523  & 26.7 & 66.2 & 28.6 & 5 \\
\hline
\textit{user\_02} & 1{,}044  & 2.9  & 8.2  & 3.5  & 1 \\
\hline
\textit{user\_03} & 12{,}172 & 34.1 & 72.9 & 31.5 & 6 \\
\hline
\textit{user\_04} & 1{,}937  & 5.4  & 12.8 & 5.5  & 2 \\
\hline
\textit{user\_05} & 1{,}196  & 3.3  & 7.7  & 3.3  & 1 \\
\hline
\textit{user\_06} & 1{,}376  & 3.9  & 9.8  & 4.2  & 2 \\
\hline
\textit{user\_07} & 1{,}241  & 3.5  & 8.5  & 3.7  & 1 \\
\hline
\textit{user\_08} & 1{,}759  & 4.9  & 11.5 & 5.0  & 1 \\
\hline
\textit{user\_09} & 1{,}784  & 5.0  & 11.5 & 5.0  & 1 \\
\hline
\textit{user\_10} & 1{,}922  & 5.4  & 11.3 & 4.9  & 1 \\
\hline
\textit{user\_11} & 1{,}774  & 5.0  & 11.0 & 4.8  & 1 \\
\hline
\textbf{Total} & \textbf{35{,}728} & \textbf{100.0} & \textbf{231.4} & \textbf{100.0} & \textbf{22} \\
\hline
\end{tabular}
\label{speaker_cont}
\end{center}
\end{table}

\section{Feature Generation}
\label{sec:impl}

In total, 28 features are extracted per 2-second analysis window preceding each candidate turn change point. Table \ref{feature_details} lists all of them with their units and observed ranges. All 28 features describe the current speaker, that is, the participant holding the turn over the 2.0\,s window preceding $t$.

\begin{table}[t]
\caption{Feature Details}
\label{feature_details}
\begin{center}
\setlength{\tabcolsep}{5pt}
\begin{tabular}{|l|l|}
\hline
\textbf{Feature} & \textbf{Unit (range)} \\
\hline
\multicolumn{2}{|l|}{\textit{Visual (9), MediaPipe Face Mesh + Py-Feat, 16\,fps}} \\
\hline
\textit{landmark\_movement}     & norm.\ coord.\ (0--100) \\
\textit{au\_intensity\_change}  & AU intensity (0--1) \\
\textit{blink\_rate}            & blinks per 2\,s (0--10) \\
\textit{gaze\_changes}          & normalized (0--1) \\
\textit{mouth\_opening\_change} & MAR (0--50) \\
\textit{head\_pose\_rotation}   & degrees (0--180) \\
\textit{translation\_changes}   & norm.\ coord.\ (0--100) \\
\textit{eyebrow\_movement}      & normalized (0--1) \\
\textit{lip\_corner\_pull}      & AU12 intensity (0--1) \\
\hline
\multicolumn{2}{|l|}{\textit{Acoustic (12), eGeMAPSv02 via openSMILE, 10\,ms frames}} \\
\hline
\textit{f0\_mean}              & Hz (80--400) \\
\textit{f0\_variation}         & ratio (0--1) \\
\textit{f0\_variation\_scaled} & ratio (0--0.5) \\
\textit{energy\_mean}          & norm.\ RMS (0--1) \\
\textit{energy\_std}           & norm.\ RMS (0--0.5) \\
\textit{energy\_rate}          & RMS\,s$^{-1}$ (unbounded) \\
\textit{onset\_rate}           & onsets/s (0--20) \\
\textit{spectral\_centroid}    & Hz (0--8000) \\
\textit{spectral\_rolloff}     & Hz (0--8000) \\
\textit{spectral\_bandwidth}   & Hz (0--8000) \\
\textit{spectral\_contrast}    & dB (0--100) \\
\textit{voiced\_ratio}         & ratio (0--1) \\
\hline
\multicolumn{2}{|l|}{\textit{Linguistic (7), from time-aligned transcripts}} \\
\hline
\textit{word\_duration}         & s (0.2--5.0) \\
\textit{syllable\_count}        & count (1--10) \\
\textit{is\_filler}             & binary $\{0,1\}$ \\
\textit{is\_question}           & binary $\{0,1\}$ \\
\textit{is\_agreement}          & binary $\{0,1\}$ \\
\textit{word\_repetition}       & binary $\{0,1\}$ \\
\textit{sentence\_completeness} & score (0--1) \\
\hline
\end{tabular}
\end{center}
\end{table}

\subsection{Visual Features}
9 features capture facial activity, gaze behavior, head motion, and expressiveness: landmark motion magnitude, action unit (AU) intensity change, blink rate, gaze direction change rate, mouth aperture variation, head rotation, head translation, eyebrow displacement, and lip corner pull. Speakers frequently display anticipatory signals such as looking away, nodding, or a drop in facial activity shortly before a turn is yielded.

\subsection{Acoustic Features}
12 features represent prosodic and spectral properties: mean $F_0$ plus its coefficient of variation and range-scaled variant; mean, standard deviation, and mean absolute per-second change of RMS energy; spectral centroid, 85\% rolloff, bandwidth, and contrast; voiced-frame ratio; and onset rate. A sudden drop in energy or a flattening pitch contour is commonly associated with a completed turn.

\subsection{Linguistic Features}
7 features capture lexical, syntactic, and discourse properties: mean word duration and mean syllables per word in the window, three binary indicators for whether the final token is a filler, whether the window contains an interrogative token, and whether it contains an affirmative token, a binary indicator of immediate word repetition, and a syntactic completeness score. Fillers and incomplete structures typically accompany continuation, whereas interrogatives and completed units mark potential transitions.

\subsection{Extraction Tools and Windowing}
Each frame is split into left and right halves corresponding to the two participants and each crop is processed independently. Landmarks, head pose, eye aspect ratio, and mouth aspect ratio are obtained with MediaPipe Face Mesh \cite{mediapipe2019} and AU intensities with Py-Feat \cite{pyfeat2023}; frame values at 16\,fps are aggregated by mean and standard deviation. Visual features are computed from the current speaker's crop only. Not using the listener's crop is a limitation, since listener gaze and nodding are established transition cues. Acoustic features are the eGeMAPSv02 set \cite{egemaps2016} extracted with openSMILE \cite{opensmile2010} at 10\,ms frames on each speaker's own channel, which also supplies $F_0$; onsets and offsets are obtained from Silero VAD \cite{silero2021} and word timings from wav2vec 2.0 forced alignment \cite{baevski2020}. Every window covers the 2.0\,s immediately preceding $t$; no information at or after $t$ enters the feature vector. No feature scaling is applied. Thresholds are sampled directly from each feature's observed range on the training partition, which keeps test-partition statistics out of the search. The GA is implemented with DEAP \cite{deap2012}.

\section{Methods}

\subsection{Problem Formulation}
The task is binary classification: given the feature vector of the two-second window preceding a conversational event, decide whether the current speaker is about to yield the turn (1) or continue (0). The framework targets rules that are interpretable and cheap enough to evaluate for use in real-time systems, and prioritizes transparency over benchmarking against black-box classifiers. Turn transitions are not assumed to follow a single cue: a speaker may yield after completing a question, after a characteristic prosodic pattern, or following a specific visual behavior. The model must therefore represent several alternative pathways, which motivates the hybrid AND--OR representation below.

\subsection{Genetic Algorithm Based Rule Optimization}

\begin{figure}[t]\centerline{\includegraphics[width=0.9\columnwidth]{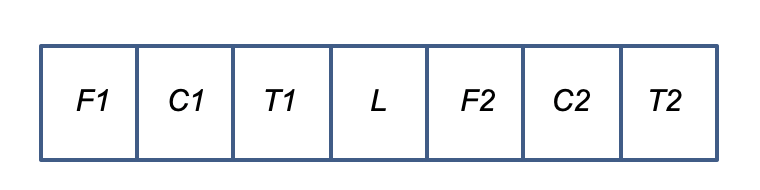}}
\caption{Simplified chromosome representation containing two features for visualization purposes.}
\label{chro}
\end{figure}

Genetic algorithms are population-based evolutionary search methods \cite{holland1975, goldberg1989} well suited to the combinatorial space of feature--threshold--operator combinations induced by rule learning. Each individual encodes a complete rule as a sequence of conditions; each condition contributes a feature, a numeric threshold, a comparison operator ($\geq$, $\leq$, $=$), and a logical operator ($\land$, $\lor$) linking it to the next, giving rules of the form \textit{if (f$_1$ c$_1$ $\theta_1$) [op] (f$_2$ c$_2$ $\theta_2$) \ldots then $y=1$, else $y=0$}. The number of conditions per chromosome is drawn uniformly between 3 and 7. Fig. \ref{chro} illustrates a two-condition chromosome, where F is feature, C comparison operator, T threshold, and L logical operator.

\begin{table}[t]
\caption{Genetic Algorithm Configuration}
\begin{center}
\setlength{\tabcolsep}{4pt}
\begin{tabular}{|l|l|}
\hline
\textbf{Parameter} & \textbf{Value} \\
\hline
Population size / max.\ generations & 500 / 900 \\
\hline
Selection                & tournament, size 5 \\
\hline
Crossover                & single-point on conditions, rate 0.85 \\
\hline
Mutation                 & rate 0.10 per offspring, 1/3 each for\\
                         & threshold, feature, logic operators \\
\hline
Elitism                  & top 5\% carried over unchanged \\
\hline
Conditions per rule      & $\mathcal{U}\{3,\ldots,7\}$ \\
\hline
Fitness                  & $F_1$ on the training partition \\
\hline
Runs per fold / seeds    & 5 / 1--5 \\
\hline
\end{tabular}
\label{ga_config}
\end{center}
\end{table}


\begin{algorithm}[t]
\caption{GA-based AND--OR rule discovery}
\label{alg:ga}
\begin{algorithmic}[1]
\REQUIRE Training dataset; Feature pool
\ENSURE Best performing rule
\STATE Initialize population: generate random rules containing 3 to 7 conditions
\FOR{generation = 1 to 900}
\STATE Calculate F1-score for all rules in the current population
\STATE Carry over the top $5\%$ of rules to the next generation \COMMENT{Elitism}
\WHILE{next generation size $< 500$}
\STATE Select two parent rules using tournament selection
\STATE Generate an offspring via crossover (85
\STATE Mutate the offspring's threshold, feature, or logic (10
\STATE Add the offspring to the next generation
\ENDWHILE
\IF{maximum F1-score has not improved for 100 generations}
\STATE \textbf{break}
\ENDIF
\ENDFOR
\RETURN Rule with the highest F1-score
\end{algorithmic}
\end{algorithm}

Table \ref{ga_config} lists the configuration and Algorithm \ref{alg:ga} the procedure. Tournament selection controls selection pressure, crossover recombines condition blocks from two parents, the three mutation operators (threshold, feature, logic) perturb numeric boundaries, replace features, and replace logical connectives, and elitism carries the current best rule into the next generation unchanged. The AND--OR encoding allows one rule to express multiple independent pathways to a transition, which a single threshold cannot represent. Candidate rules are scored by $F_1$ on the training partition: optimizing precision alone yields conservative rules that miss true transitions, optimizing recall alone produces a high false-positive rate, and $F_1$ balances the two. A prediction counts as a true positive when it falls within 0.5\,s of an annotated turn change, which prevents temporal imprecision in the reference timestamps from being charged as an error; unmatched predictions are false positives and unmatched references false negatives. The same tolerance is applied to all baselines in Table \ref{model_compare}.

\subsection{Cross-Validation Strategy}
\label{sec:cv}
A 5-fold cross-validation is applied with partitioning at the conversation rather than the sample level, ensuring that samples from one conversation do not appear in both training and test. Because 11 conversations do not divide evenly into five parts, the folds hold three, two, two, two, and two conversations respectively. For each fold, four folds train and one tests; GA optimization runs on the training data only and the discovered rule is evaluated on the unseen test conversations. Each fold serves as test partition exactly once, and precision, recall, and $F_1$ are averaged across folds.

Conversation-level partitioning eliminates conversation-specific leakage, while speaker-specific leakage remains. Because every conversation contains user\_01 or user\_03 (Section II), a fully speaker-disjoint 5-fold split cannot be realized here and the same speakers necessarily occur in training and test folds. A fully speaker-disjoint estimate would require a leave-one-speaker-out protocol over the nine non-recurring speakers, which is left to future work; the reported figures should be read as an upper bound on generalization to unseen speakers.

\section{Results}

Table \ref{generation_performance} reports the evolution of test-fold performance and Table \ref{model_compare} compares the final rule against baselines; values are means over the 5 folds. The silence baselines predict a turn change at $t$ whenever the pause immediately preceding $t$, taken from the Silero VAD offsets, exceeds $\tau \in \{1.0, 2.0, 3.0\}$\,s; no other information is used. The three reference baselines follow analytically from the 1{,}750:3{,}500 class ratio.

Because the class ratio is 1:2 (1{,}750 positives to 3{,}500 negatives across the corpus, and preserved within every fold), always predicting a turn change attains precision $33.3$, recall $100.0$, $F_1 = 50.0$. The always-positive baseline is the primary reference point used below. Against it the GA rule improves $F_1$ by 7.0 points and raises precision from the $33.3\%$ base rate to $46.1\%$ at $74.6\%$ recall. All three silence thresholds fall below it. This follows from the corpus composition described in Section \ref{sec:labelling}: non-speech is rare (12.4\%) and 81.4\% of transitions involve overlap, which leaves pause duration weakly informative here. The gap to the silence baselines is therefore not interpreted as evidence of high predictive accuracy, and given the 11 conversations in the corpus, no claim of statistical significance is made for the 7.0-point margin over this baseline.

\begin{table}[t]
\caption{Evolution of Model Performance Across Generations (5-fold averages)}
\begin{center}
\setlength{\tabcolsep}{4pt}
\begin{tabular}{|c|c|c|c|}
\hline
\textbf{Generations} & \textbf{Precision} & \textbf{Recall} & \textbf{$F_1$} \\
\hline
0   & 32.4 & 49.3 & 39.1 \\
\hline
100 & 38.6 & 60.4 & 47.1 \\
\hline
300 & 45.1 & 66.0 & 53.6 \\
\hline
500 & 45.9 & 71.2 & 55.8 \\
\hline
700 & 46.0 & 73.6 & 56.6 \\
\hline
900 & 46.1 & 74.6 & 57.0 \\
\hline
\end{tabular}
\label{generation_performance}
\end{center}
\end{table}

\begin{table}[t]
\caption{Performance Results}
\begin{center}
\setlength{\tabcolsep}{3pt}
\begin{tabular}{|l|c|c|c|}
\hline
\textbf{Model} & \textbf{Precision} & \textbf{Recall} & \textbf{$F_1$} \\
\hline
Always predict turn change & 33.3 & 100.0 & 50.0 \\
\hline
Random, $p = 0.5$          & 33.3 & 50.0  & 40.0 \\
\hline
Stratified random          & 33.3 & 33.3  & 33.3 \\
\hline
Silence threshold $1.0$\,s      & 10.5 & 40.3  & 16.7 \\
\hline
Silence threshold $2.0$\,s      & 20.4 & 35.0  & 25.8 \\
\hline
Silence threshold $3.0$\,s      & 22.2 & 10.7  & 14.4 \\
\hline
GA rule (proposed)                     & 46.1 & 74.6  & \textbf{57.0} \\
\hline
\end{tabular}
\label{model_compare}
\end{center}
\end{table}

After the evaluation phase, the GA was re-trained on the entire dataset to derive one interpretable rule, given in \eqref{eq:rule} and reported for qualitative discussion. Thresholds are given in the units listed in Table \ref{feature_details}:

\begin{equation}
\label{eq:rule}
\begin{split}
(\text{word\_duration} \geq 0.60\,\mathrm{s} \land \text{energy\_rate} \geq 1.00) \\
\lor (\text{gaze\_changes} \geq 0.35 \land \text{f0\_mean} \geq 120\,\mathrm{Hz}) \\
\lor (\text{is\_filler} = 1 \land \text{word\_duration} \geq 0.80\,\mathrm{s})
\end{split}
\end{equation}

In the rule, longer word duration combined with energy change acts as a prosodic closing cue, gaze shifts paired with mean $F_0$ form a visual-acoustic transition signal, and fillers, although insufficient alone, become informative when prolonged. Only 5 of the 28 features are selected and the visual modality contributes one condition; the three modalities do not contribute equally in the final rule.




\section{Conclusion}

This study introduced a multimodal Turkish turn-taking corpus of 4.23 hours of synchronized video, per-speaker audio, and time-stamped transcripts with 1{,}750 filtered turn-change events, together with a rule optimization procedure for predicting turn transitions from these signals. Instead of a black-box classifier, prediction is formulated as binary classification over interpretable rules evolved with a hybrid AND--OR representation, on the premise that prosodic, visual, and linguistic mechanisms can independently trigger a transfer. The resulting rule reaches $F_1 = 57.0\%$, above both the silence thresholds and the always-positive baseline of $F_1 = 50.0$, and it can be inspected directly. The margin over that baseline is modest. Future work will use a balanced, non-hub recording design with speaker-normalized acoustic features, compare against Random Forest, XGBoost, and transformer-based classifiers, and integrate the rules into real-time LLM-based agents.

\end{document}